\pdfoutput=1
\documentclass{article}
\usepackage[utf8]{inputenc}

\usepackage[english]{babel}
\usepackage[utf8]{inputenc}
\usepackage[T1]{fontenc}

\usepackage{enumitem}

\usepackage[a4paper,top=3cm,bottom=2cm,left=3cm,right=3cm,marginparwidth=1.75cm]{geometry}

\usepackage{amsmath}
\usepackage{verbatim}
\usepackage{amssymb}
\usepackage{mathrsfs}
\usepackage{stmaryrd}
\usepackage{graphicx}
\usepackage[singlelinecheck=false]{caption} 
\usepackage[square,comma,numbers,sort&compress]{natbib}
\usepackage{bbold}
\usepackage[colorinlistoftodos]{todonotes}
\usepackage[colorlinks=true, allcolors=blue]{hyperref}
\usepackage[french,onelanguage]{algorithm2e}
\usepackage{multirow,tabularx}
\usepackage{diagbox}
\usepackage{lscape}
\hypersetup{colorlinks,%
citecolor=blue,%
filecolor=black,%
linkcolor=black,%
urlcolor=black
}
\usepackage[super]{nth}
\usepackage{pdfpages}
\newcommand{\distas}[1]{\mathbin{\overset{#1}{\kern\z@\sim}}}%

\usepackage[justification=centering]{caption}
\usepackage{subfig}
\usepackage{array}
\makeatletter
\newcommand{\thickhline}{%
    \noalign {\ifnum 0=`}\fi \hrule height 0.8pt
    \futurelet \reserved@a \@xhline
}

\begin{document}

%TITRE + AUTEURS + AFFILIATION
\title{\textbf{Spread-gram: A spreading-activation schema of network structural learning}}
\author{Jie Bai\textsuperscript{1}, Linjing Li\textsuperscript{1}, Daniel Zeng\textsuperscript{1} \\ \\
\textsuperscript{1}The State Key Laboratory of Management and Control for Complex Systems,\\
Institute of Automation, Chinese Academy of Sciences,\\
Beijing, China\\ 
\textbf{\{baijie2013, linjing.li, dajun.zeng\}@ia.ac.cn}
}

\date{}
\maketitle
%\keywords{machine learning;deep learning;embedded systems;security;adversarial machine learning;neural network quantization}

%ABSTRACT
\section*{Abstract}
\label{Abstract}
Network representation learning has exploded recently. 
However, existing studies usually reconstruct networks as sequences or matrices, which may cause information bias or sparsity problem during model training.
Inspired by a cognitive model of human memory, we propose a network representation learning scheme.
In this scheme, we learn node embeddings by adjusting the proximity of nodes traversing the spreading structure of the network. 
Our proposed method shows a significant improvement in multiple analysis tasks based on various real-world networks, ranging from semantic networks to protein interaction networks, international trade networks, human behavior networks, etc. 
In particular, our model can effectively discover the hierarchical structures in networks.
The well-organized model training speeds up the convergence to only a small number of iterations, and the training time is linear with respect to the edge numbers. 

\textbf{Keywords: }representation learning, network analysis, Spreading Activation, cognitive psychology, node embeddings

%INTRODUCTION
\section{Introduction}
\label{Introduction}

Modern society contains innumerable networks \cite{easley2010networks}, such as social network, the World Wide Web, knowledge graphs and various information networks constructed by human behaviours. 
%Analyze these networks is necessary to understand the world better. 
Network representation learning, also known as structure representation \cite{battaglia2018relational} or network embedding \cite{tang2015line}, is an unique building block of understanding and analyzing the networks \cite{bengio2013representation}. 
This technique map the networks into a latent space where the structural semantic information condenses, i.e., node embeddings, which can also be seen as the dimensional reduction for networks. 
Similar research questions also include latent space model \cite{sarkar2006dynamic} and graph embedding \cite{yan2007graph}, who try to make the structural semantics of the network computable. 
 
Recent network representation learning studies, including random-walk-based methods \cite{perozzi2014deepwalk, grover2016node2vec, dong2017metapath2vec, ivanov2018anonymous} and neural-network-based methods \cite{cao2016deep, liu2017semantic}, develop with the advance of deep learning techniques and have made major process in many specific network analysis tasks.
% which has shown their success in various fields such as computer vision, speech recognition and natural language processing \citep{bengio2013representation,mikolov2013efficient}. 
However, these deep learning techniques are mostly designed for text or images originally. Unlike text and images whose element assignment are fixed (for example, a sentence is a sequence of words, an image is a matrix of pixels), the topological structure of the network is uncertain, i.e., for each node in a network, we cannot fix the number of its neighbors in advance. 
The uncertainty of the topological structure in networks leads to the particular requirements of network representation learning, where the common sequential input scheme will brining in bias, and the matrix input scheme will suffer from the data sparsity. These drawbacks can easily damage the globalized high-order structure of networks \citep{benson2016higher}. 

\begin{figure*}[t]
\centering
\includegraphics[width=12cm]{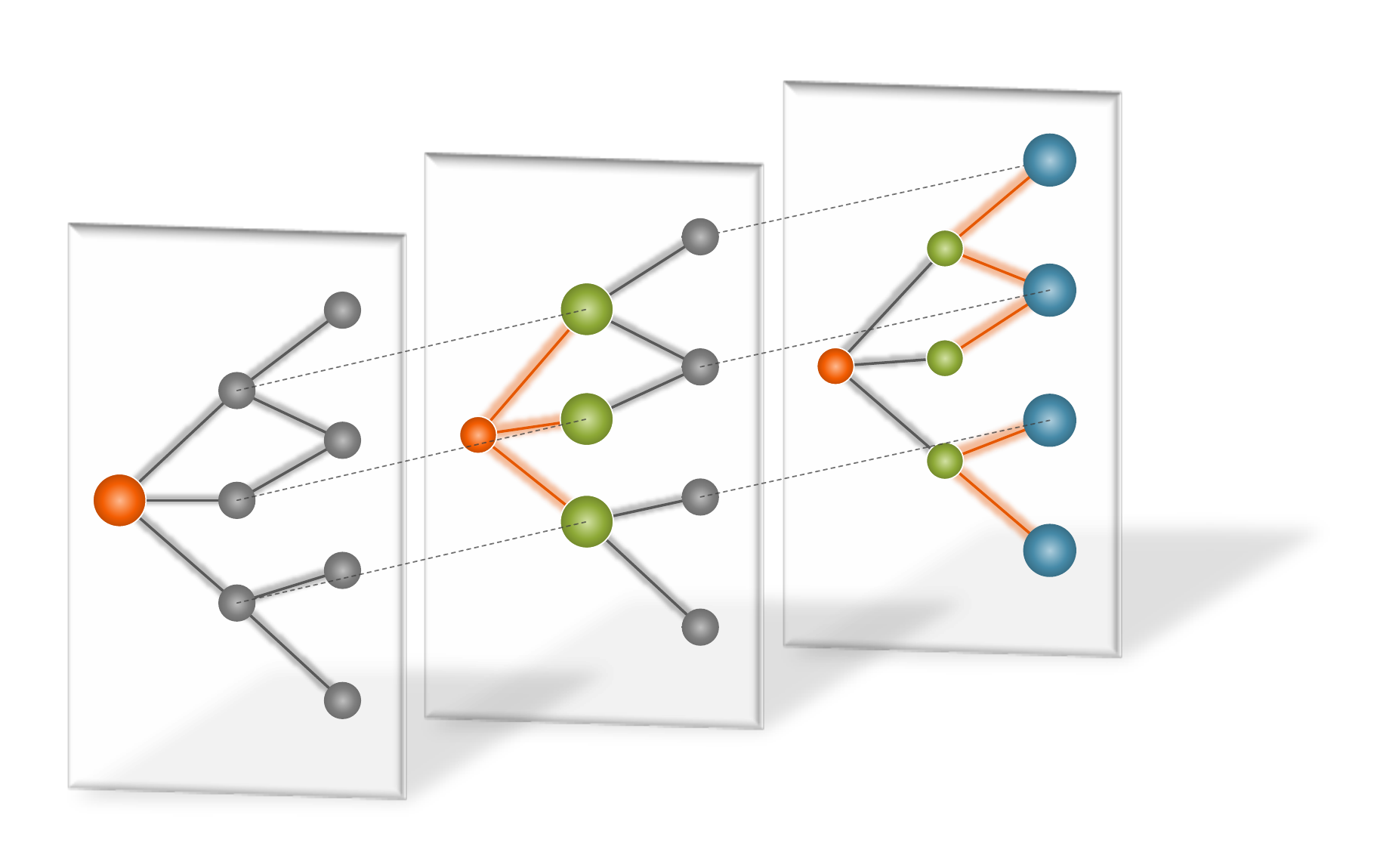}
\caption{Illustration of the spreading-activation mechanism. Different colors indicates the order of the nodes being searched.}
\label{fig:illustration}
\end{figure*}

On the other side, it is widely received that the cognitive mechanisms of human intelligence are skilled at representing structure and reasoning about relations based on networks \cite{battaglia2018relational}. Many cognitive theories believed that human knowledge system is stored as networks (graph), and high-order cognitive activities, like learning, reasoning and problem solving, are operations based on these networks \cite{anderson1982acquisition, griffiths2010probabilistic, jonker2018neural}. Among those various theories, the spreading-activation theories are a major branch in cognitive psychology which study concentratedly on the interactions of entities in the memory network of human cognition. They describes the dynamical memory searching process happened in human brain. Figure \ref{fig:illustration} illustrates the common spreading-activation mechanism. The spreading-activation mechanism overcome the challenge we discussed before, i.e., it fulfills the uncertaintity of network topological structure, and handles both global structure and local correlations simultaneously. The first spreading activation theory was proposed to implement computer simulations of memory search \cite{collins1975spreading}. Since proposed, it has been applied extensively in semantic network, information retrieval \cite{crestani1997application, preece1982spreading, shoval1981expert, cohen1987information} and social network analysis \cite{ziegler2004spreading, dasgupta2008social, yang2010study, wang2018entagrec++}. However, to our knowledge, how to apply spreading activation theories in the field of network representation learning has not been well studied.

We propose spread-gram, a spreading-activation schema of network representation learning. 
Spread-gram is inspired by the ACT spreading-activation theory \cite{anderson1983spreading}, which addresses the dynamical retrieval issue in human brain when complex memory activity happen. 
Our intuition is that spreading-activation process in human cognitive nature can help building better network analysis schema. In spread-gram, the node embeddings are learned following the spreading structure of network, each time the proximity of node pairs along the spreading paths will be compared and adjusted. More specifically, given a source node, the activation procedure will spread to all of the targeted nodes that connected with the source node. And then these targeted nodes will become new source nodes and spread the activation procedure to all of their neighbors. The spreading will terminate until all nodes in the connected graph have been activated once. On one hand, this learning schema will make sure that the nodes have all been selected and updated along the network structure, so the non-linear structure is permitted and the input bias can be avoid. On the other hand, the well-organized activation and learning will speed-up the convergence of model training.

\begin{figure*}[!htbp]
\centering
\includegraphics[width=12cm]{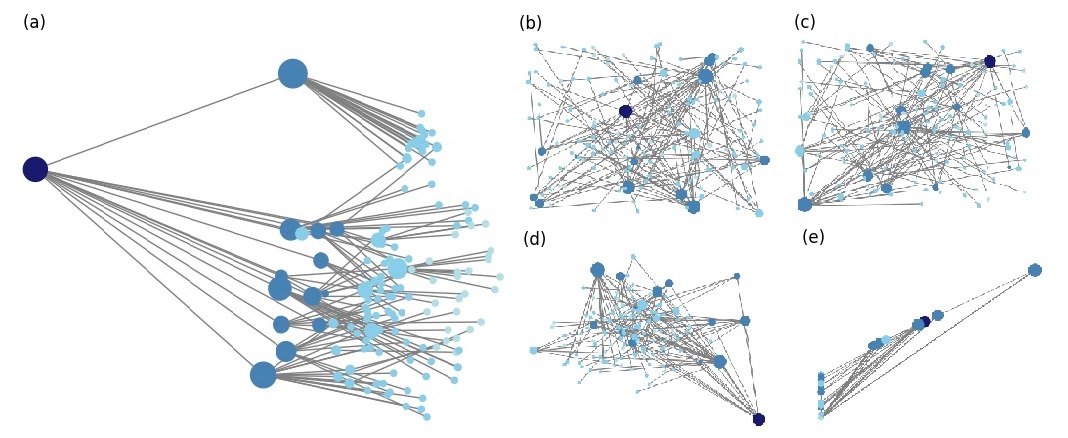}
\caption{Network representation learning for discovering hierarchical structure of the networks. The network in use is the taxonomy of the  Wikipedia entries, where the hierarchies are labeled by different colors. (a) is the 2-D visualization of our proposed model, (b)-(e) are that of some typical baseline models, i.e., DeepWalk, Node2Vec, TransE and GCN seperately. More details are described in the ``node embedding visualization'' part.}\label{fig:hierarchy}
\end{figure*}

We discuss the availability and the approach that apply the ACT spreading activation theory to network representation learning. Then the network searching schema and node representation updating rules are constructed accordingly. Finally, we propose network representation learning methods for both homogeneous and heterogeneous separately. 
Through the experiment, we found that spread-gram is good at preserve both local and global correlations of the networks. Its distinct ability is on discovering the hierarchical structure of the networks, which can hardly be captured by previous works. The related experimental results are shown in Figure \ref{fig:hierarchy}. Moreover, spread-gram succeeds against multiple network analysis tasks in a wide range of real-world networks.

Compared with existing network representation learning methods, the superiority of Spread-gram lies in:

\begin{itemize}
\item	The model is designed specifically for the uncertainty of network topological structure, avoiding the information bias or sparsity problem that may happened in other models;
\item   It integrates both global and local structure information of networks, which shows a significant advance in learning and representing the hierarchical structure;
\item	The well-organized model training speeds up the convergence to only a small number of iterations, and the training time grows linear in the number of the edge numbers.
\end{itemize}

%Method
\section{Method Design}
\label{Method}
\subsection{Problem Formulation}
Existing literatures \cite{han2012mining} usually define a network as $G = \left( {V,E} \right)$, where $V$ is the collection of the nodes and $E$ is the collection of the edges. For the heterogeneous networks, there are type-mapping functions $\phi :V \to A$ and $\psi :E \to R$, which map $V$ and $E$ to the node-type collection $A$ and edge-type collection $R$, separately.

Under this context, network representation learning can be defined as mapping $G = \left( {V,E} \right)$ to an implicit vector space ${R^d}$, i.e, learning a mapping ${\alpha _G}:V \to {R^d}$, such that for any $v \in V$, ${\alpha _G}\left( v \right)$ preserves the structural semantics of $v$ as much as possible. Here $d \ll R$ is the dimensionality of the implicit vector space. Given that the object of this research is the network $G$, we will abbreviate ${\alpha _G}\left( v \right)$ as $\alpha \left( v \right)$ in the following descriptions.

Based on the definitions above, we need to first define the metric which can preserve the semantic information of the network. Existing literatures tried to explain and reconstruct network semantics from various perspectives, such as inner product of node vectors for distance representation \cite{perozzi2014deepwalk, tang2015line}, Laplacian Eigenmaps for global neighborhood relationships \cite{belkin2002laplacian, wang2016structural}. These studies usually construct loss functions by comparing vector similarities with actual relationships of node pairs. Our work integrate this idea with the structural characteristics of the ACT spreading activation formula. The ACT spreading activation formula is represented as \cite{anderson1983spreading}:

\begin{eqnarray} \label{eqn:1}
a(y) = \sum\limits_x {f(x,y) \cdot a} (x) + c(y),
\end{eqnarray}

where $a(x)$ and $a(y)$ represent the activation rate of $x$ and $y$ separately, $f(x,y)$ represents the correlation between $x$ and $y$, and $c(y)$ represents the baseline activation of $y$. Note that $a(x)$ and $a(y)$ serve equivalent posts in this equation, i.e., they can be replaced by each other in the further transmissions.

Inspired by the ACT spreading activation formula, we generalized $a(x)$ and $a(y)$ from scalar to vector representations, and take $c(y)$ as the former state of $a(y)$. Then equation \ref{eqn:1} can be transformed to:

\begin{eqnarray} \label{eqn:2}
{\alpha ^{t + 1}}(y) = \sum\limits_{x \in N(y)} {{f^{t + 1}}(x,y) \cdot {\alpha ^t}(x) + {\alpha ^t}(y)}.
\end{eqnarray}

The equation \ref{eqn:2} means that the representation of \textit{y} is adjusted according to its surrounding nodes. Here $N(y)$ is the collection of y’s neighbors, ${f^{t + 1}}(x,y)$ represents the correlation that obtained through ${a^t}(x)$ and ${a^t}(y)$, i.e., 

\begin{eqnarray} \label{eqn:3}
{f^{t + 1}}(x,y) = f({a^t}(x),{a^t}(y))
\end{eqnarray}

Now we need to construct the objective function which will make the update function fulfills equation \ref{eqn:2}. We will first consider the situations in homogeneous networks, then the model inference will be generalized to heterogeneous networks.

\subsection{Homogeneous Networks}

In heterogeneous networks, $x$ and $y$ are of the same type, so $a(x)$ and $a(y)$ share the same semantic space. Considering the structure of equation \ref{eqn:2}, we can construct a log-linear model as the objective function. Then maximum likelihood estimation can be used to obtain the gradient for model updating, i.e., to obtain ${f^{t + 1}}(x,y)$.

Let

\begin{eqnarray} \label{eqn:4}
P(u|x,y) = {{{{\left\{ {\exp [a(x)a(y)]} \right\}}^u}} \over {1 + \exp [a(x)a(y)]}},
\end{eqnarray}

Here \textit{u} represents the association of $x$ and $y$. If $x \in N(y)$ then $u = 1$, otherwise $u = 0$. $a(x)a(y) \in ( - \infty , + \infty )$ is the inner product of $a(x)$ and $a(y)$, it can reflect the associations of $x$ and $y$ in the context of the vector space. $P(u|x,y) \in (0,1)$ is the likelihood function of $a(x)a(y)$, which can be used to define the consistence between the associations of node pairs and that of the corresponding node vectors.

Define a sigmoid function:

\begin{eqnarray} \label{eqn:5}
\sigma (x,y) = {{\exp [a(x)a(y)]} \over {1 + \exp [a(x)a(y)]}},
\end{eqnarray}

then equation \ref{eqn:4} can be transformed to:

\begin{eqnarray} \label{eqn:6}
P(u|x,y) = \sigma {(x,y)^u}{\left[ {1 - \sigma (x,y)} \right]^{1 - u}};
\end{eqnarray}

For any node pair $\left( {x,y} \right)$, equation \ref{eqn:6} holds. As a result, the log likelihood of the network is:

\begin{eqnarray} \label{eqn:7}
	\begin{split}
  	L(G) &= \log \left[ {\prod\limits_{y \in V} {\prod\limits_{x \in V} {P(u|x,y)} } } \right]         \\
    &= \sum\limits_{y \in V} {\sum\limits_{x \in V} {\left[ {u\log \sigma (x,y) + (1 - u)\log (1 - \sigma (x,y))} \right]}}.
	\end{split}
\end{eqnarray}

The gradient of $a(x)$ and $a(y)$ under $L(G)$ should be:

\begin{eqnarray} \label{eqn:8}
	\begin{split}
  & {{\partial L} \over {\partial a(y)}} = \sum\limits_{x \in N(y)} {a(x)\left[ {u - \sigma (x,y)} \right]},  \\
  & {{\partial L} \over {\partial a(x)}} = \sum\limits_{y \in N(x)} {a(y)\left[ {u - \sigma (x,y)} \right]} {\kern 1pt} {\kern 1pt} {\kern 1pt}.
	\end{split}
\end{eqnarray}

We can see that the format of ${{\partial L} \over {\partial a(x)}}$ and ${{\partial L} \over {\partial a(y)}}$ are the same, although $x$ and $y$ serves as different roles in the objective function. It means that $x$ and $y$ are interconvertible in the general model training process. Putting equation \ref{eqn:2} and equation \ref{eqn:8} together, we can get:

\begin{eqnarray} \label{eqn:9}
{f^{t + 1}}(x,y) = \left[ {u - \sigma (x,y)} \right].
\end{eqnarray}

Then the node vector updating function can be represented as:

\begin{eqnarray} \label{eqn:10}
{a^{t + 1}}(y) = \sum\limits_{x \in N(y)} {{a^t}(x)\left[ {u - \sigma (x,y)} \right] + {a^t}(y)}.
\end{eqnarray}

Equation \ref{eqn:10} implies that, the node embedding updating equation is identified with the ACT spreading activation formula, when the objective function is defined by log-linear model, the node-pair association is represented through the inner product of their embeddings, and the parameters are evaluated with the maximum-likelihood estimation.

\subsection{Heterogeneous Networks}

According to the definition of network above, in a heterogeneous network there are more than one type of nodes or edges. In this paper we only consider the heterogeneous networks who have more than one type of nodes. Assuming that each node type has its specific semantic space, the problem is how to integrate different semantic spaces together to perform representation learning. We address this problem by designing a benchmark space and a transforming mechanism among different semantic spaces.

Let ${{\mathbb{R}}_{0}}$ be the benchmark space, ${{W}_{t}}$ be the mapping matrix from ${{\mathbb{R}}_{t}}$ to ${{\mathbb{R}}_{0}}$. For any $v\in V$, its type is $\phi (v)$, the correlated vector representations of $ v $ in space ${{\mathbb{R}}_{0}}$ is:

\begin{eqnarray} \label{eqn:11}
{{\alpha }_{0}}(v)={{W}_{\phi (v)}}\alpha (v).
\end{eqnarray}

Then for any $x$ and $y$ in the network, their correlation in the benchmark space should be:

\begin{eqnarray} \label{eqn:12}
{{a}_{0}}(x) {{a}_{0}}(y)=a{{(x)}^{T}}{{W}_{\phi (x)}}^{T}{{W}_{\phi (y)}}a(y).
\end{eqnarray}

According to the analysis of homogeneous networks, let

\begin{eqnarray} \label{eqn:13}
	\begin{split}
    {{\sigma }_{0}}(x,y)&=\frac{\exp [{{a}_{0}}(x)\centerdot {{a}_{0}}(y)]}{1+\exp [{{a}_{0}}(x)\centerdot {{a}_{0}}(y)]} \\ 
 &=\frac{\exp [a{{(x)}^{T}}{{W}_{\phi (x)}}^{T}{{W}_{\phi (y)}}a(y)]}{1+\exp [a{{(x)}^{T}}{{W}_{\phi (x)}}^{T}{{W}_{\phi (y)}}a(y)]}, \\ 
	\end{split}
\end{eqnarray}

So the log-linear model of heterogeneous networks is:

\begin{eqnarray} \label{eqn:14}
	\begin{split}
 L(G)&=\log \left[ \prod\limits_{y\in V}{\prod\limits_{x\in V}{P(u|x,y)}} \right] \\ 
 &=\sum\limits_{y\in V}{\sum\limits_{x\in V}{\left[ u\log {{\sigma }_{0}}(x,y)+(1-u)\log (1-{{\sigma }_{0}}(x,y)) \right]}}. \\ 
	\end{split}
\end{eqnarray}

The gradient of $a(x)$ and $a(y)$ should be:

\begin{eqnarray} \label{eqn:15}
	\begin{aligned}
  & \frac{\partial L}{\partial a(y)}=\sum\limits_{x\in N(y)}{{{W}_{\phi (y)}}^{T}{{W}_{\phi (x)}}a(x)\left[ u-{{\sigma }_{0}}(x,y) \right]}, \\ 
  & \frac{\partial L}{\partial a(x)}=\sum\limits_{y\in N(x)}{{{W}_{\phi (x)}}^{T}{{W}_{\phi (y)}}a(y)\left[ u-{{\sigma }_{0}}(x,y) \right]}. \\ 
	\end{aligned}
\end{eqnarray}

The gradient of ${{W}_{\phi (x)}}$ and ${{W}_{\phi (y)}}$ should be:

\begin{eqnarray} \label{eqn:16}
	\begin{aligned}
  & \frac{\partial L}{\partial {{W}_{\phi (y)}}}=\sum\limits_{y\in V}{\sum\limits_{x\in V}{{{W}_{\phi (x)}}\alpha (x)\alpha {{(y)}^{T}}\left[ l-{{\sigma }_{0}}(x,y) \right]}}, \\ 
  & \frac{\partial L}{\partial {{W}_{\phi (x)}}}=\sum\limits_{x\in V}{\sum\limits_{y\in V}{{{W}_{\phi (y)}}\alpha (y)\alpha {{(x)}^{T}}\left[ l-{{\sigma }_{0}}(x,y) \right]}}. \\ 
	\end{aligned}
\end{eqnarray}

\section{Node Search Strategy and Model Training}

\subsection{Node Search}

\begin{figure}[!htbp]
\centering
\subfloat [“A-B, A-D” schema]{\includegraphics[width=4cm]{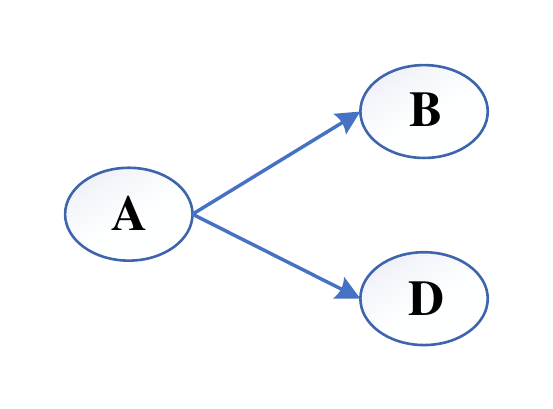}
\label{fig:ab-ad}}
\hfil
\subfloat [“A-B, C-D” schema]{\includegraphics[width=4cm]{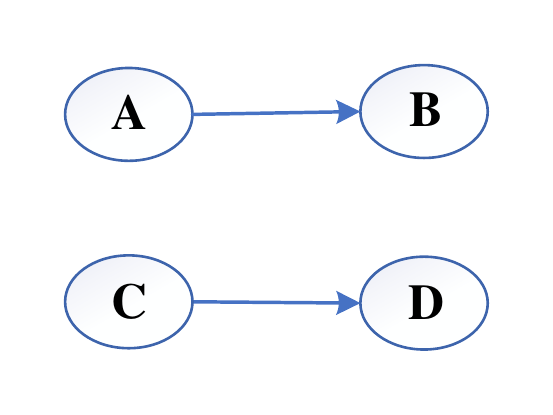}
\label{fig:ab-cd}}
\caption{Two typical spreading schema.}
\label{fig:spreading-schema}
\end{figure}

The ACT spreading-activation theory proposed two typical spreading schema \cite{anderson1983spreading}, which is called ``A-B, A-D'' and ``A-B, C-D'' schemas. ``A-B, A-D'' means the spreading process starting from a specific node and spread to multiple targets; ``A-B, C-D'' means the spreading proceed concurrently from multiple origins to multiple targets. Figure \ref{fig:spreading-schema} illustrates these schemas.

\begin{figure}[!t]
\centering
\includegraphics[width=12cm]{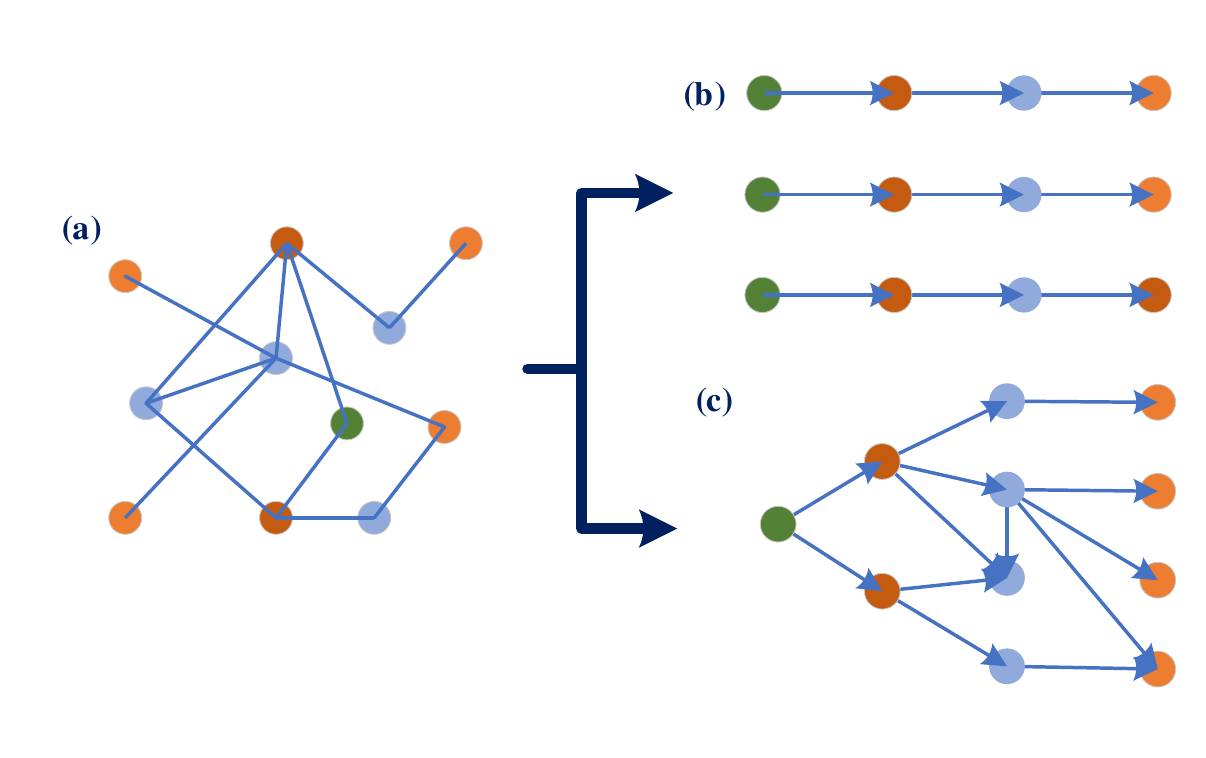}
\caption{Node searching of spread-gram in a connected network.}\label{fig:network-search}
\end{figure}

The strategy of node searching and updating in spread-gram integrates ``A-B, A-D'' and ``A-B, C-D'' schemas. A node searching example using spread-gram is shown in Figure \ref{fig:network-search}. It selects a node randomly as the source, spread to its neighbors (``A-B, A-D''); then taking these neighbors as sources, and spread to their neighbors. Note that from here the spreading progress proceed in parallel (``A-B, C-D''). This node searching strategy ensure the spreading and updating reaching every node in the network. Besides, the ordered updating is more efficient than random updating, since every source node is updated before spreading to its neighbors. For a network constituted by multiple connected one, the spread-gram node searching conducted on these connected networks respectively.

Table \ref{table:algorithm} describes a realization of spread-gram node searching algorithm.

\begin{table}[!t]
\renewcommand{\arraystretch}{1.3}
\caption{A realization of spread-gram node searching algorithm}
\label{table:algorithm}
\centering
\begin{tabular}{l}
\hline
Input: node set $V$, edge set $E$\\
Output: Order of the activated nodes, a sequence $Y$\\
\hline
\textbf{Algorithm: spread-gram node searching}\\
    ~~~~Initialize ${Y} =$ an empty list, ${Q} = $ an empty queue, $N =$ an empty map\\
    ~~~~for $v$ in $V$\\
    ~~~~~~~~$N\left( v \right)\leftarrow $ node set of u for each $\left( u,v \right)$ in $ E $\\
    ~~~~$v\leftarrow$ sample form $V$ ~~~~~~~~~~~~~~~~~~~~~~~~~~~~~~~~~~~~~~~~~~~~~~~~~~~~~~~~~~~~~~~~~~~~~~~~\textbf{step *}\\
    ~~~~append $ v $ to $ Q $\\
    ~~~~while $ Y $ isn’t empty, do\\
    ~~~~~~~~if $ Q $ isn’t empty, then\\
    ~~~~~~~~~~~~$v\leftarrow$ Pop a node form $Q$\\
    ~~~~~~~~~~~~if $v$ in $V$, then\\
    ~~~~~~~~~~~~~~~~append $ v $ to $ Y $, remove $ v $ from $ V $\\
    ~~~~~~~~~~~~~~~~append all of $ u $ in $ N\left( v \right) $ to $ Q $\\
    ~~~~~~~~else\\
    ~~~~~~~~~~~~return to \textbf{step *}\\
    ~~~~end for\\
    ~~~~return ${\bf{Y}}$\\
\hline
\end{tabular}
\end{table}

\subsection{Model Training}

In this part we will discuss the spread-gram model training. First is the training samples construction. According to the objective functions (equation \ref{eqn:7} and equation \ref{eqn:14}), each $y\in V$ need to compute the likelihood with all of the other nodes in the network. However, it should be noticed that the nonadjacent node pairs in a network usually far more than the adjacent ones. So the origin objective functions could lead to the huge computational consumption and unbalanced models potentially. 
We adopt the negative sampling strategy \cite{mikolov2013distributed} to construct the training samples and solve the above problem. More specifically, each time we choose an adjacent node pair $(x,y)$ as a training sample, a fixed number of nonadjacent node pairs $(x',y)$ related to y will also be included, where $x'$ is sampled randomly from the network.

Let $k\in N*$ denotes the coefficient of the negative sampling, $N(y)$ is the neighbor set of $y$, and we will choose $k\times |N(y)|$ nodes from the nonadjacent nodes in $V$ for each $y$ as negative samples, along with $N(y)$ adjacent nodes as positive samples. Therefore the training samples $N'(y)$ corresponding to $y$ should be:

\textbf{\begin{eqnarray} \label{eqn:17}
N'\left( y \right)=N\left( y \right)\bigcup \left\{ {{x}_{1}},...,{{x}_{k\times \left| N\left( y \right) \right|}}|x\notin N\left( y \right),x\ne y \right\}
\end{eqnarray}}

Based on equation \ref{eqn:17}, the objective functions need to be adjusted. For example, equation \ref{eqn:7} is reformulated as:

\begin{eqnarray} \label{eqn:18}
	\begin{split}
  L(G)&=\log \left[ \prod\limits_{y\in V}{\prod\limits_{x\in {N}'(y)}{P(l|x,y)}} \right] \\ 
 &=\sum\limits_{y\in V}{\sum\limits_{x\in {N}'(y)}{\left[ l\log \sigma (x,y)+(1-l)\log (1-\sigma (x,y)) \right]}}.
	\end{split}
\end{eqnarray}

For $y$ and $x\in {N}'(y)$, the update function should be:

\begin{eqnarray} \label{eqn:19}
	\begin{split}
  {{\alpha }^{t+1}}(y)&=\gamma \sum\limits_{x\in {N}'(y)}{{{\alpha }^{t}}(x)\left[ l-\sigma (x,y) \right]+{{\alpha }^{t}}(y)}, \\ 
 {{\alpha }^{t+1}}(x)&=\gamma {{\alpha }^{t}}(y)\left[ l-\sigma (x,y) \right]+{{\alpha }^{t}}(x).
	\end{split}
\end{eqnarray}

%Complexity
\section{Complexity Analysis}
\label{Complexity}
A spread-gram model iteration is constructed by two components: node searching and parameter updating. We discuss the complexity of the algorithm by analyzing these two components.

In the node searching process, the algorithm goes through the network with the spreading mechanism and generates a node list to decide the order of nodes’ activation. We should preserve a dynamical node query $Q$ and an ordered list $Y$. When the searching process beginning, we continuously pop the node from $Q$ and check if it is activated (exist in $Y$). If not, append the node to $Y$ and push all of its neighbors to $Q$. Where there is an edge, there should be a node being added to the $Q$ (some nodes may be added repeatedly). Meanwhile, each node in $Q$ should be checked the existence in $Y$. In each iteration, there will be $|E|$ nodes adding to $Q$. If there is a hash set to record the activated nodes, the existence checking for a single node will cost $O(1)$ time. Therefore the time complexity of node searching process is $O(|E|)\times O(1)=O(|E|)$.

For the parameter updating process, we should discuss the situations in homogeneous networks and heterogeneous networks, separately. In homogeneous networks, for each node $y$ we should first construct a collection of training nodes ${N}'(y)$. Then for each $x\in {N}'(y)$, compute its association with \textit{y} and update the vectors of \textit{x} and \textit{y} according to equation \ref{eqn:10}. Although the size of ${N}'(y)$ varies according to y, the total amount of nodes in all ${N}'(y)$s is computable, i.e., $\sum\limits_{y\in V}{|}{N}'(y)|=2|E|(k+1)$, where \textit{k} is the parameter of negative sampling. The time consumption of vector updating lies in the calculating of inner product and summation of the vectors, whose time complexity are both $O(d)$ where \textit{d} is the dimension of the node embeddings. As a result, the time complexity of parameter updating in homogeneous networks is $O(2k|E|)\times O(d)=O(kd|E|)$.

In heterogeneous networks, we should further consider the involving of mapping matrix. For a single pair of nodes, the time consumption of calculating their inner product in the benchmark space is $O({{d}^{2}})+O({{d}^{2}})+O(d)=O({{d}^{2}})$. Therefore, the time consumption of infer the gradient and updating a pair of nodes should be $O({{d}^{2}})+O({{d}^{2}})+O({{d}^{2}})+O(d)=O({{d}^{2}})$, and the node embedding training process in an iteration will take $O(k|E|)\times O({{d}^{2}})=O(k{{d}^{2}}|E|)$ time. The mapping matrixes are updated at the end of an iteration, which needs to consider all of the training node pairs and involves matrix operations. The time consumption of updating single matrix is $O(k|E|)\times O({{d}^{2}})=O(k{{d}^{2}}|E|)$. Because there are $|T|$ mapping matrixes that need to be updated, so the total time consumption of training mapping matrix should be $O(k{{d}^{2}}|E||T|)$.

%EXPERIMENTS
\section{Experiments}
\label{Experiments}
\subsection{Toy case study}
\begin{figure}[t]
\centering
\subfloat [A simple connected network]{\includegraphics[width=12cm]{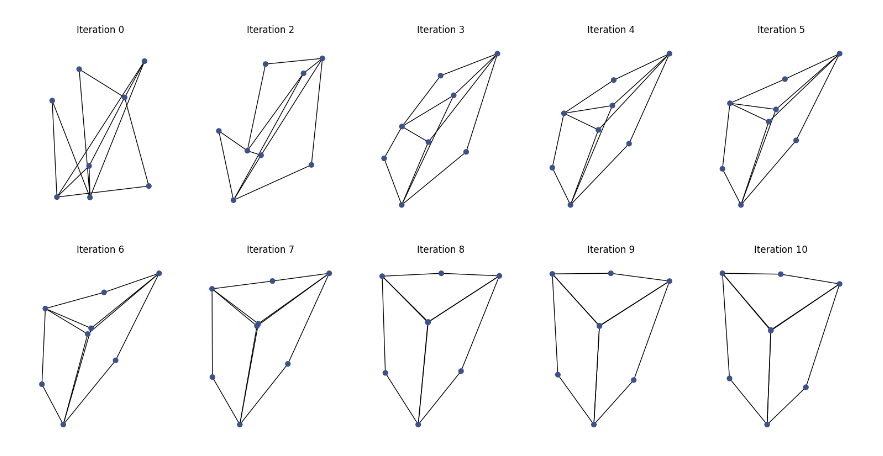}
\label{fig:toy1}}
\hfil
\subfloat [A complex network constructed by a few connected graphs]{\includegraphics[width=12cm]{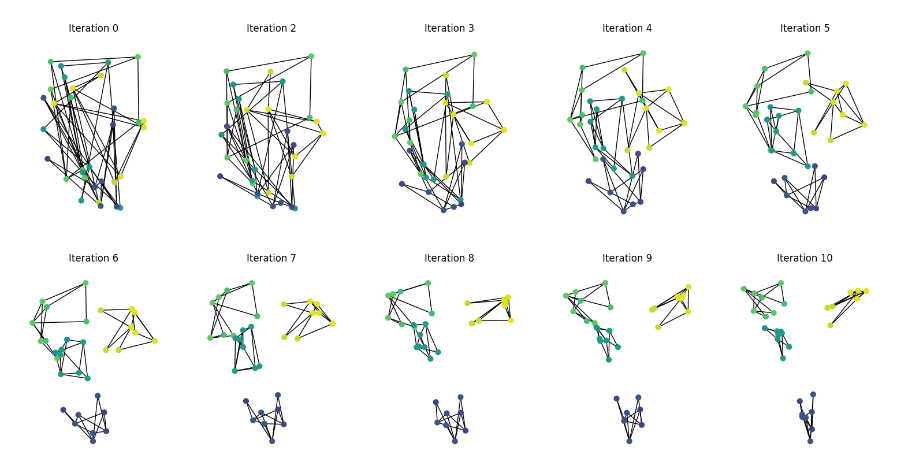}
\label{fig:toy2}}
\caption{Spread-gram model iterations on two toy cases. 
The case (a) is a connected network with 8 nodes. 
During the spread-gram iterations, the structure of the network was learned and represented better and better gradually, and two nodes that have same neighbors shared the same location. 
The case (b) is a network containing multiple connected graphs. 
The spread-gram model identified the connected graphs and distributed them with different locations. 
All of the learning and representation is converged within just a few iterations.}
\label{fig:toy}
\end{figure}

We first observed the iteration process of spread-gram through its running on toy case networks. 
We used two typical networks (a simple connected network and a simple complex network constitute by multiple connect networks), learned 2-dimensional node embeddings on each network, and visualized the network through the node embeddings (random initialized).
The result is shown in Figure \ref{fig:toy}.

\subsection{Experimental Settings for Real-World Networks}

Experiments are conducted for various real-world networks, including undirected networks, directed networks and heterogeneous networks. Multiple tasks were conducted to evaluate the performance of the models quantitatively and qualitatively. 

\subsubsection{Methods}

The methods in comparison are as follows:

\begin{itemize}
\item \textbf{Spread-gram} is the network representation learning model proposed by this paper. The modeling strategy (homogeneous or heterogeneous) is decided by the nature of the network. The maximum number of iteration is set as 30. The dimension of the node embeddings $d=128$.
\item \textbf{DeepWalk} \cite{perozzi2014deepwalk} is a classical random-walk based representation learning method for homogeneous networks. Because that the length of node sequences in DeepWalk is identical with the number of iterations in spread-gram, we set it as 30. The window weigh of training node embeddings is 8, the dimension of node embeddings is 128 as well.
\item \textbf{Node2vec} \cite{grover2016node2vec} is a network representation learning method which combines the depth-first search and breadth-first search and can be seen as a generalization of DeepWalk. For the hyper-parameter p and q in the model, we set them as 0.5 and 2 separately, which makes the random-walk tends to be breadth-first searching. The other parameter settings of node2vec is the same as that of DeepWalk.
\item \textbf{Metapath2vec++} \cite{dong2017metapath2vec} is a state-of-the-art random-walk based heterogeneous network representation learning method following a predefined metapath schema. We adopted ``A-P-A'' as the meta-path of the model. The other parameters are the same with that in DeepWalk.
\item \textbf{PTE} \cite{tang2015pte} is a semi-supervised network representation learning method which is superior in preserving semantic correlations among nodes. PTE obeys a sequential learning schema during the model training. We set the window size of the training model as 5 and the batch size as 300.
\item \textbf{GCN} \cite{kipf2016semi} is a typical neural network based network representation learning method. During the training of the GCN models, the number of iterations is set as 30, and the dimension of node embeddings is set as 128, which are same as the settings in Spread-gram.
\item \textbf{TransE} \cite{bordes2013translating} is a typical knowledge graph representation learning model. Knowledge graph is a specific type of network, so we chose a representative method for learning knowledge graphs for comparison. For the parameter settings, we sampled 10,000 batches for model training, and each batch has 100 triples. A triples is a minimal unit of the knowledge graph, which is constructed by a head node, a tail node and an edge. For better training of the model, we ranked the network to make sure that there is no circle on top of the triples.
\end{itemize}

\subsubsection{Datasets}

We chose five datasets including both homogeneous and heterogeneous networks. The datasets come from various fields, such as bioinformatics, economics and human behaviors. For all of the datasets, there are group information for the nodes, so supervised classification experiments can be conducted conveniently. The details about the datasets are as follows.

\begin{itemize}
\item \textbf{WITS\footnote{https://wits.worldbank.org/Default.aspx?lang=en}} is an international trading analysis tool provided by the World Bank. We downloaded the country-wide trading data from WITS which contain 233 countries/regions along with the top frequent import and export country during the last few years for each country/region. By taking the countries as nodes and the import/export relationships as edges, we constructed a directed homogeneous network with 233 nodes and 4301 edges. The continents the countries/regions belong were set as node categories.
\item \textbf{Wiki} is a semantic correlation dataset collected by West and Leskovec \cite{west2012human}. This dataset is constructed based on the Wikipedia websites, which has 4,592 entries and 119,882 hyperlinks. We built the homogeneous network through taking the entries as nodes and the hyperlinks between entries as edges. Besides, there is a hierarchical taxonomy to locate the entries. By choosing the first-stage categories in the taxonomy to classify the entries, we could classify the entries into 15 categories, including science, history, art, and so on. 
\item \textbf{DIP\footnote{https://dip.doe-mbi.ucla.edu/dip/Main.cgi}}  is a protein interaction database collected by UCLA. It contains a large amount of experimentally determined interactions between different proteins. This dataset has 28,255 proteins and 76,881 interactions. We built the homogeneous network through taking the proteins as nodes, the interactions as edges and the proteins’ types as categories.
\item \textbf{DBLP\footnote{https://dblp.uni-trier.de/faq/How+can+I+download+the+whole+dblp+dataset}}is a bibliographical dataset provided by the DBLP, a website collecting and managing bibliographical information from the field of computer science. This dataset is dynamically increased, which has accumulated 216,636 literatures published by 619,626 authors until our downloading. If taking the authors and the literatures as two types of nodes, tanking the publishing behavior as the edges, we can get a heterogeneous network with 836,262 nodes and 1,605,633 edges. The group information in this dataset is the publication venue, such as the name of the journal or conference.
\item \textbf{Amazon} is a production review dataset collected from Amazon online shopping website \cite{leskovec2007dynamics}. Based on the assumption that a review is corresponding to a buying behavior, we constructed an online shopping network constituted by products and consumers. This network contains 986,934 nodes and 2,329,915 edges. The categories of the products can be used as the group information.
\end{itemize}

The following of this part will be organized according to the different tasks, which include (1) link prediction, (2) node classification, (3) node visualization, and (4) iteration analysis. To unify the networks, we set the edge weight of the networks as 1 uniformly. When learning node embeddings for the different networks using different models, the learning rate is chosen from the range of 0.01 to 0.05. For all of the classification tasks (link prediction, node classification), we divided the dataset into 70\% training set and 30\% testing set.

\subsection{Link Prediction}

The link prediction task is used to examine the consistency of the proximity between a pair of nodes and that between their corresponding node embeddings. It evaluates the models’ ability to preserve the semantic correlation among nodes. In this task we trained a binary classifier, whose input is the difference of the embeddings of a specific node pair, and output is a binary value to indicate whether this pair of nodes are connected by an edge. During data construction for the link prediction, for each node we randomly chose one of its neighbors as a positive sample, and one node who is not its neighbor as a negative sample. For the heterogeneous networks, a pair of nodes should be selected from different types, and the node embeddings learned from the spread-gram method needs to be mapped to the benchmark space before calculating their relative locations. Metapath2vec++ is used in heterogeneous networks specially, so there will be no result of applying Metapath2vec++ to homogeneous networks, i.e., Wiki, WITS and DIP.

We adopt SVM with RBF core as the classifier. The settings about the classifier is: penalty rate $C$=500, parameter of the RBF core $\gamma =1/128$. Average accuracy is used as the metric to evaluate the models’ performance. The experimental results are shown in Table \ref{table:link-prediction}.

\begin{table}%[tbhp]
\centering
\caption{Link prediction results}
\label{table:link-prediction}
\begin{tabular}{lrrrrr}
\hline
Dataset & WITS & Wiki & DIP & DBLP & Amazon\\
\hline
spread-gram & \textbf{0.756} & \textbf{0.864} & \textbf{0.891} & \textbf{0.989} & \textbf{0.930} \\
DeepWalk & 0.565 & 0.601 & 0.851 & 0.664 & 0.595 \\
Node2vec & 0.580 & 0.650 & 0.847 & 0.667 & 0.599 \\
TransE & 0.519 & 0.538 & 0.589 & 0.501 & 0.557 \\
PTE & 0.674 & 0.770 & 0.621 & 0.647 & 0.708 \\
GCN & 0.611 & 0.628 & 0.512 & 0.696 & 0.707 \\
Metapath2vec++ & - & - & - & 0.690 & 0.718 \\
\hline
\end{tabular}
%\addtabletext{nomenclature for the TSs refers to the numbered species in the table.}
\end{table}

From the experimental results in Table \ref{table:link-prediction}, we can see that our proposed model spread-gram outperforms the other models significantly in the link prediction task. It is not surprising because spread-gram is good at learning the correlations between nodes through global input and spreading-learning strategy.

\subsection{Node Classification}

The node classification task evaluates the node embeddings with respect to discriminations on different categories. For the 5 datasets in use, the number of categories are various. We ranked the categories in each dataset based on the number of entries, and selected the top 7-15 categories for each dataset to conduct node classification. Specifically, for the heterogeneous networks, there are more than one type of nodes; for some specific node types, there is more than one category that each node should belong to, such as the ``authors'' nodes in DBLP, and the ``consumers'' nodes in Amazon. As a result, multi-label is needed for these node types.

\begin{figure}[!htbp]
\centering
\subfloat [ ``Countries'' nodes in WITS network]{\includegraphics[width=7cm]{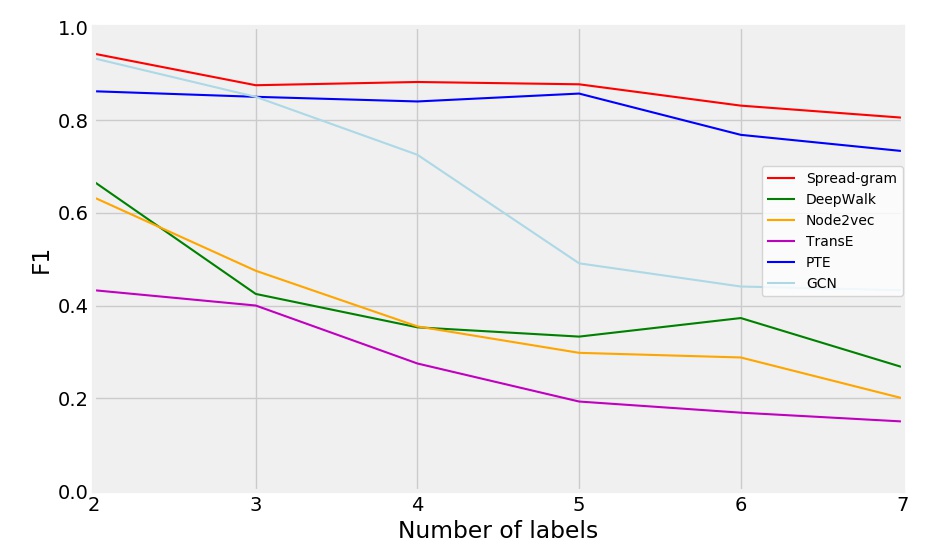}
\label{fig:node-homo-a}}
\hfil
\subfloat [``Entries'' nodes in Wiki network]{\includegraphics[width=7cm]{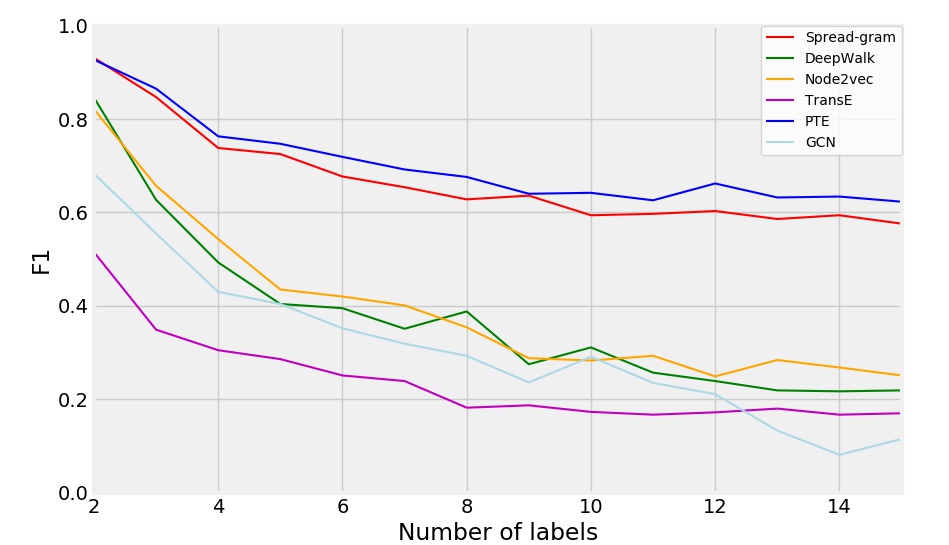}
\label{fig:node-homo-b}}
\hfil
\subfloat [``Proteins'' nodes in DIP network]{\includegraphics[width=7cm]{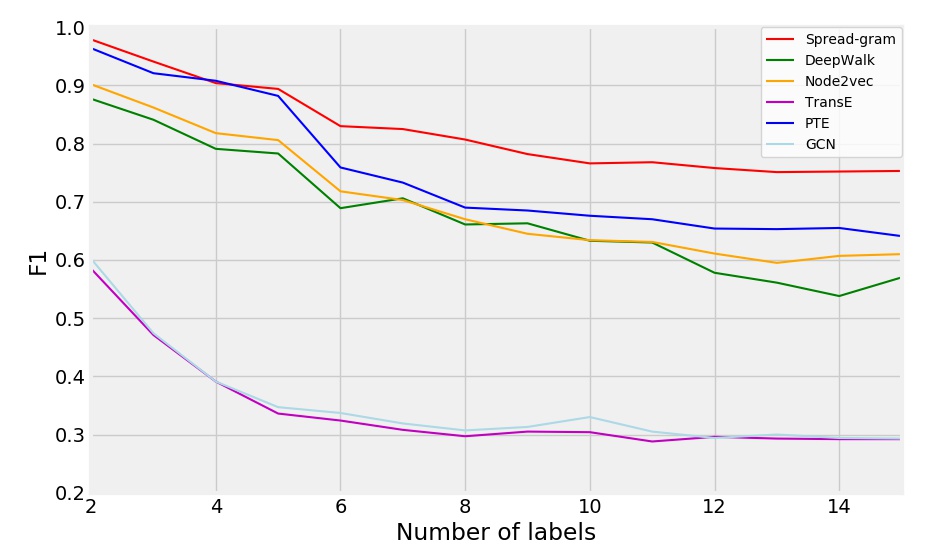}
\label{fig:node-homo-c}}
\caption{Node classification results on homogeneous networks.}
\label{fig:node-homo}
\end{figure}

\begin{figure}[!htbp]
\centering
\subfloat [``Authors'' nodes in DBLP network ]{\includegraphics[width=7cm]{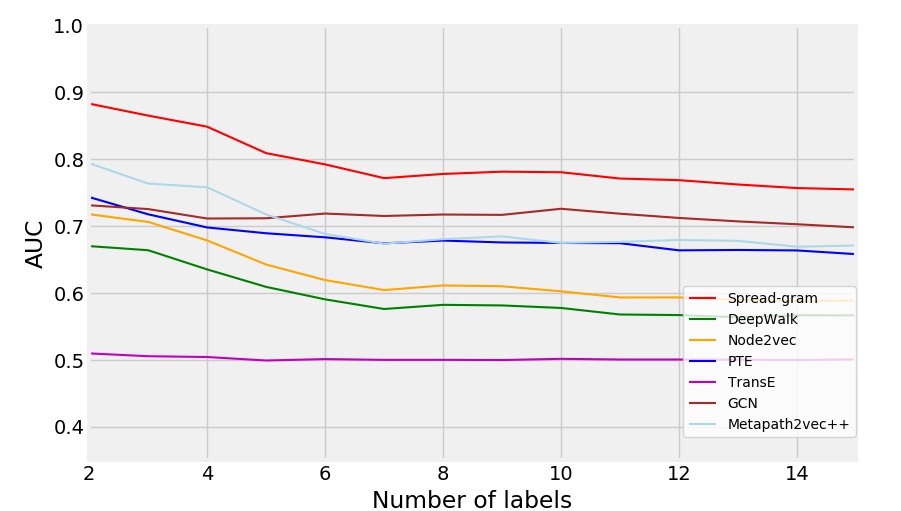}
\label{fig:node-hetero-a}}
\subfloat [``Publications'' nodes in DBLP network]{\includegraphics[width=7cm]{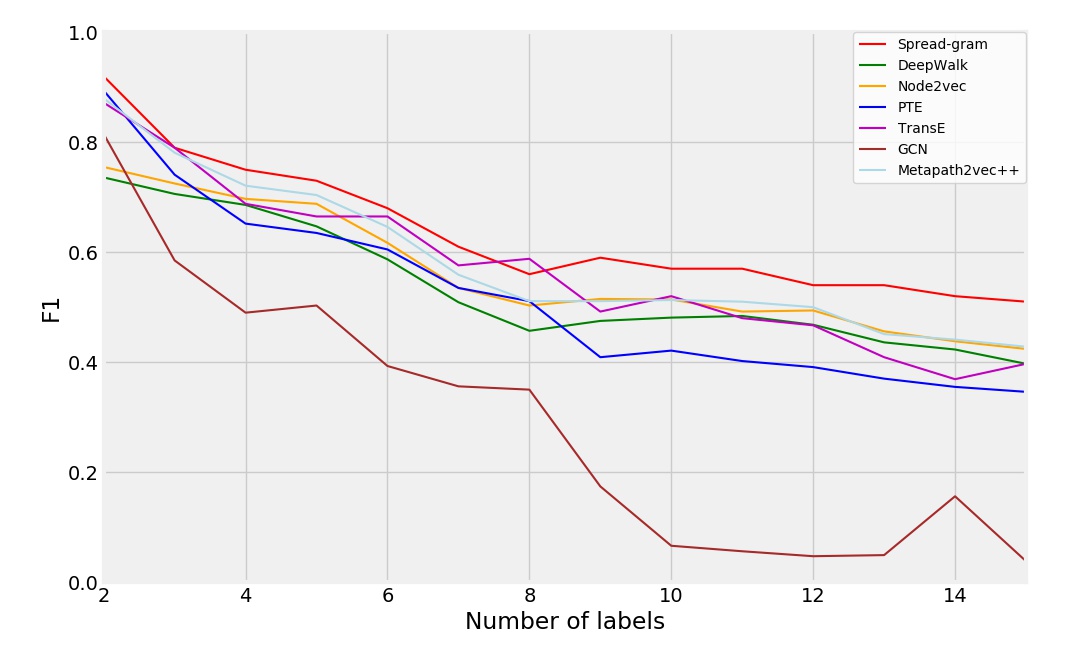}
\label{fig:node-hetero-b}}
\hfil
\subfloat [``Consumers'' nodes in Amazon network]{\includegraphics[width=7cm]{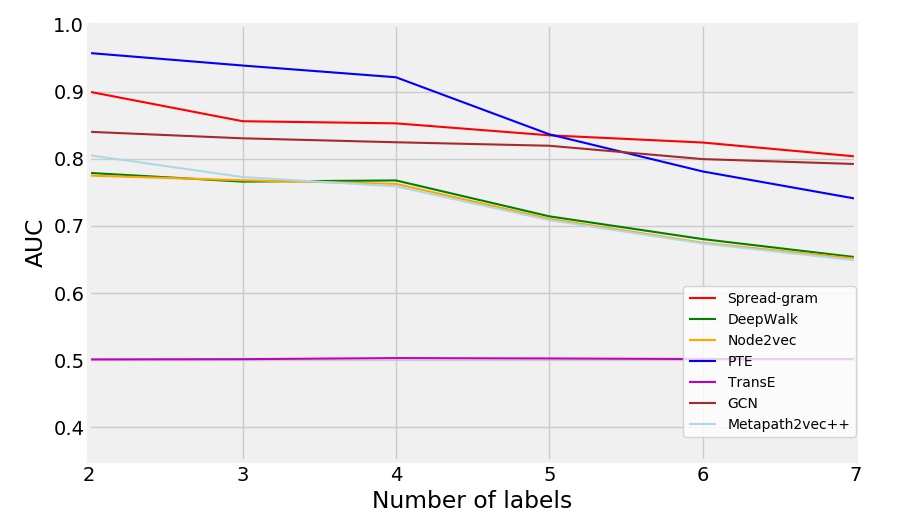}
\label{fig:node-hetero-c}}
\subfloat [``Products'' nodes in Amazon network]{\includegraphics[width=7cm]{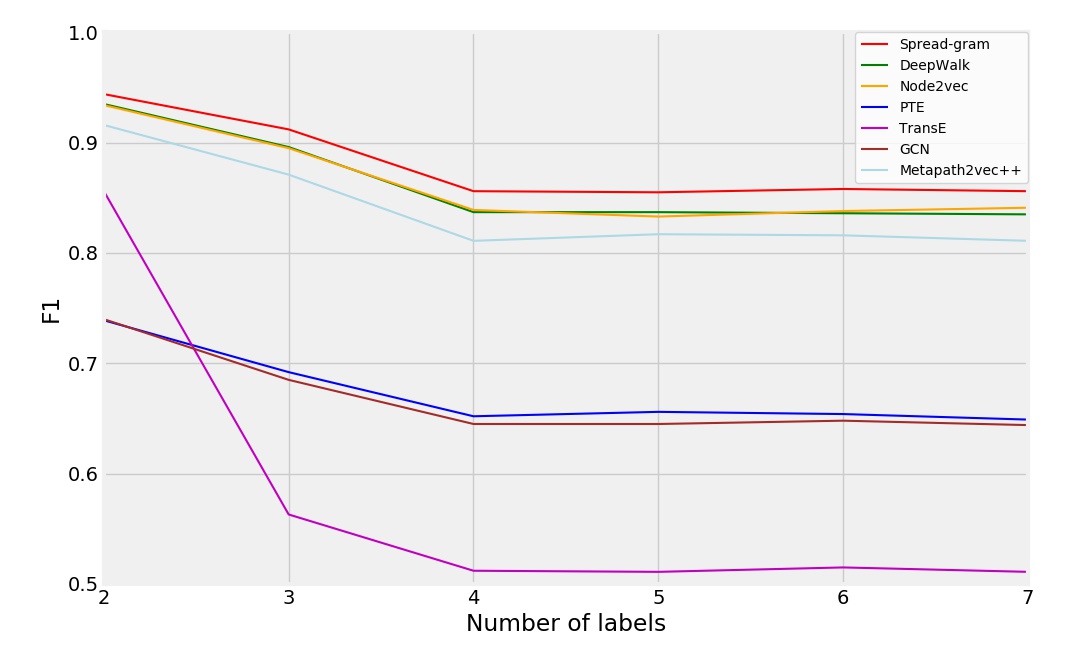}
\label{fig:node-hetero-d}}
\caption{Node classification results on heterogeneous networks.}
\label{fig:node-hetero}
\end{figure}

For the multi-label node classification experiments, we adopted the decision tree as the classifier with AUC as the metric; for the other classification experiments, we used the SVM as the classifier with average F1 score as the metric. To get more details about the performance of the models, we conducted multiple classifications for each model on each dataset, by increasing the amount of the labels gradually. The experimental results are shown as follows, where Figure \ref{fig:node-homo} shows the results on the homogeneous networks, and Figure \ref{fig:node-hetero} shows the results on the heterogeneous networks.

Figure \ref{fig:node-homo} and Figure \ref{fig:node-hetero} show that spread-gram succeed in the node classification tasks generally, i.e., for most cases, spread-gram performs better and more stable than the baseline models. We can also see that some specific method like PTE produces better results in some specific cases. However, it is reasonable because PTE is a semi-supervised method which has leveraged the label information during model training. In contrast, in spread-gram only node-node correlations is used, and the global structure like node category belonging is learned effectively. It indicates that spread-gram is able to preserve both local and global information for the networks.

\subsection{Node Embedding Visualization}

In the node embedding visualization task, we plot the nodes so as to qualitatively evaluate the distribution of node embeddings learned by different methods. Among the 5 networks, the nodes in Wiki contain abundant semantic information, so we chose it to compare the network structure and spatial correlations of nodes generated by the models.

\begin{figure}[!htbp]
\centering
\subfloat [Spread-gram]{\includegraphics[width=7cm]{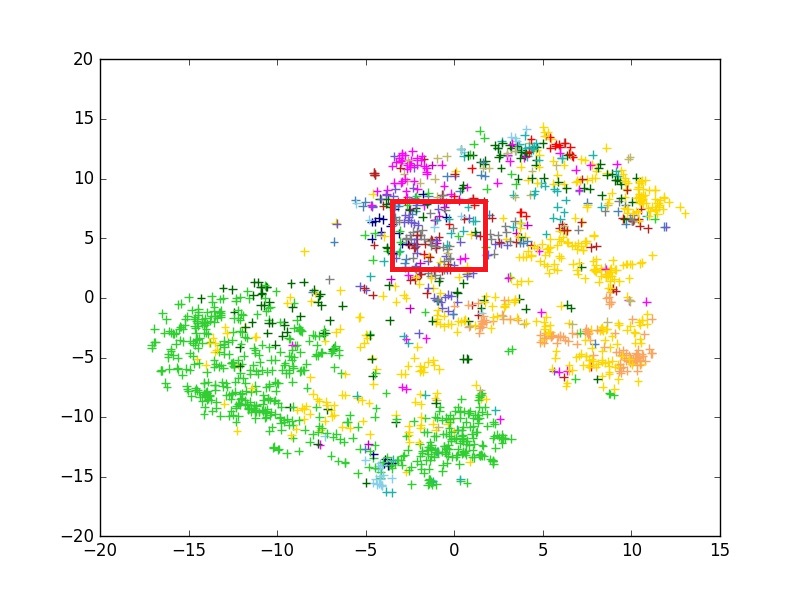}
\label{fig:visualization-a}}
\subfloat [DeepWalk]{\includegraphics[width=7cm]{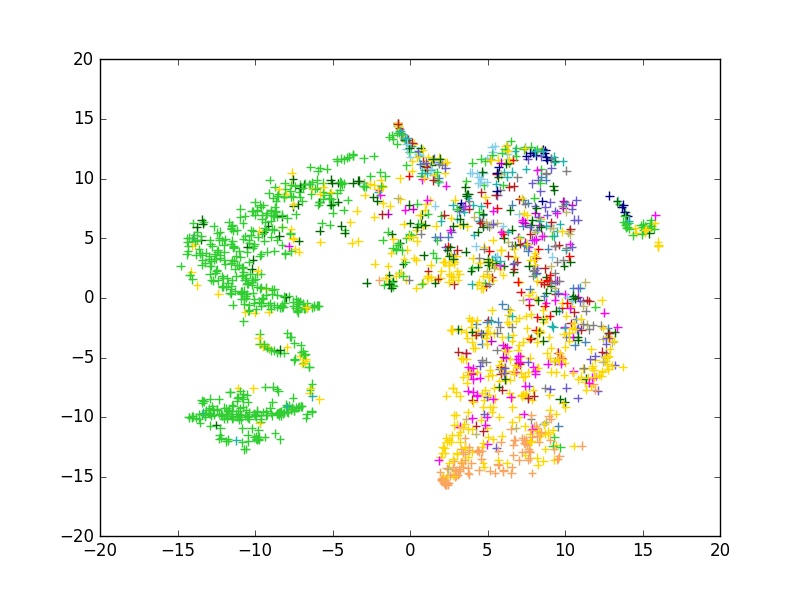}
\label{fig:visualization-b}}
\hfil
\subfloat [Node2vec]{\includegraphics[width=7cm]{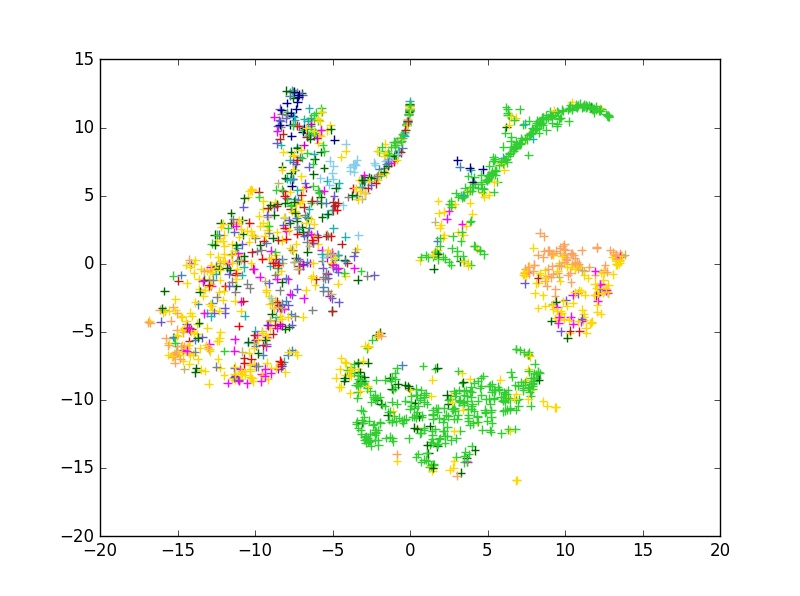}
\label{fig:visualization-c}}
\subfloat [TransE]{\includegraphics[width=7cm]{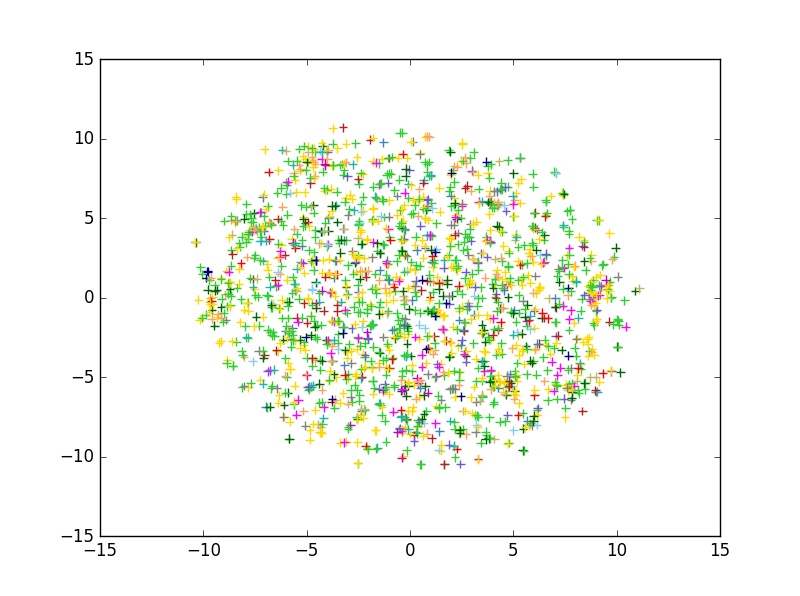}
\label{fig:visualization-d}}
\hfil
\subfloat [PTE]{\includegraphics[width=7cm]{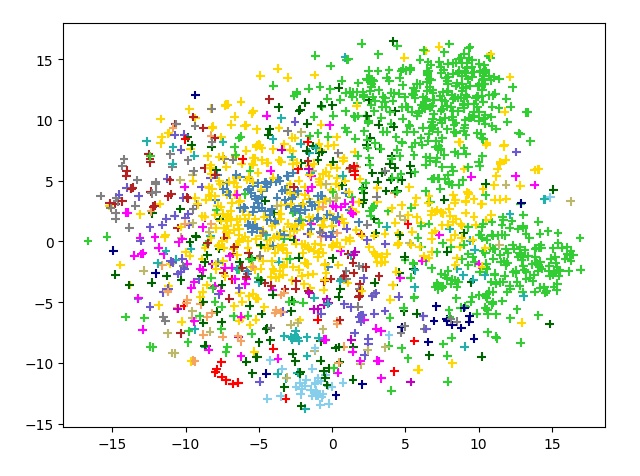}
\label{fig:visualization-e}}
\subfloat [GCN]{\includegraphics[width=7cm]{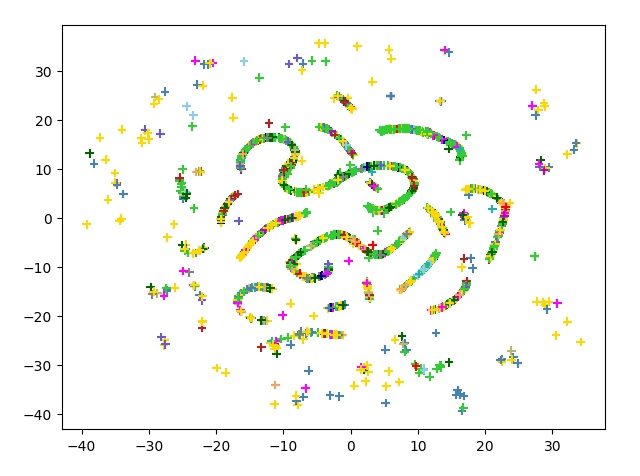}
\label{fig:visualization-f}}
\caption{Node embeddings distribution on a 2-diminsional plane.}
\label{fig:visualization}
\end{figure}

The origin node embeddings are 128-dimensional vectors. For the convenience of visualization, we condensed the node embeddings to two dimensional vectors through T-SNE \cite{maaten2008visualizing}, a manifold-learning based dimensional reduction method. The vectors were then plotted on a two-dimensional plane, and were labeled by different colors to separate their categories. To present more clearly, we selected and showed the nodes which refer to independent and general concepts. For example, the node of ``Library'' was plotted but ``Library\_of\_Alexandira'' wasn’t. The experimental results are shown in Figure \ref{fig:visualization}.

\begin{figure}[!htbp]
\centering
\includegraphics[width=12cm]{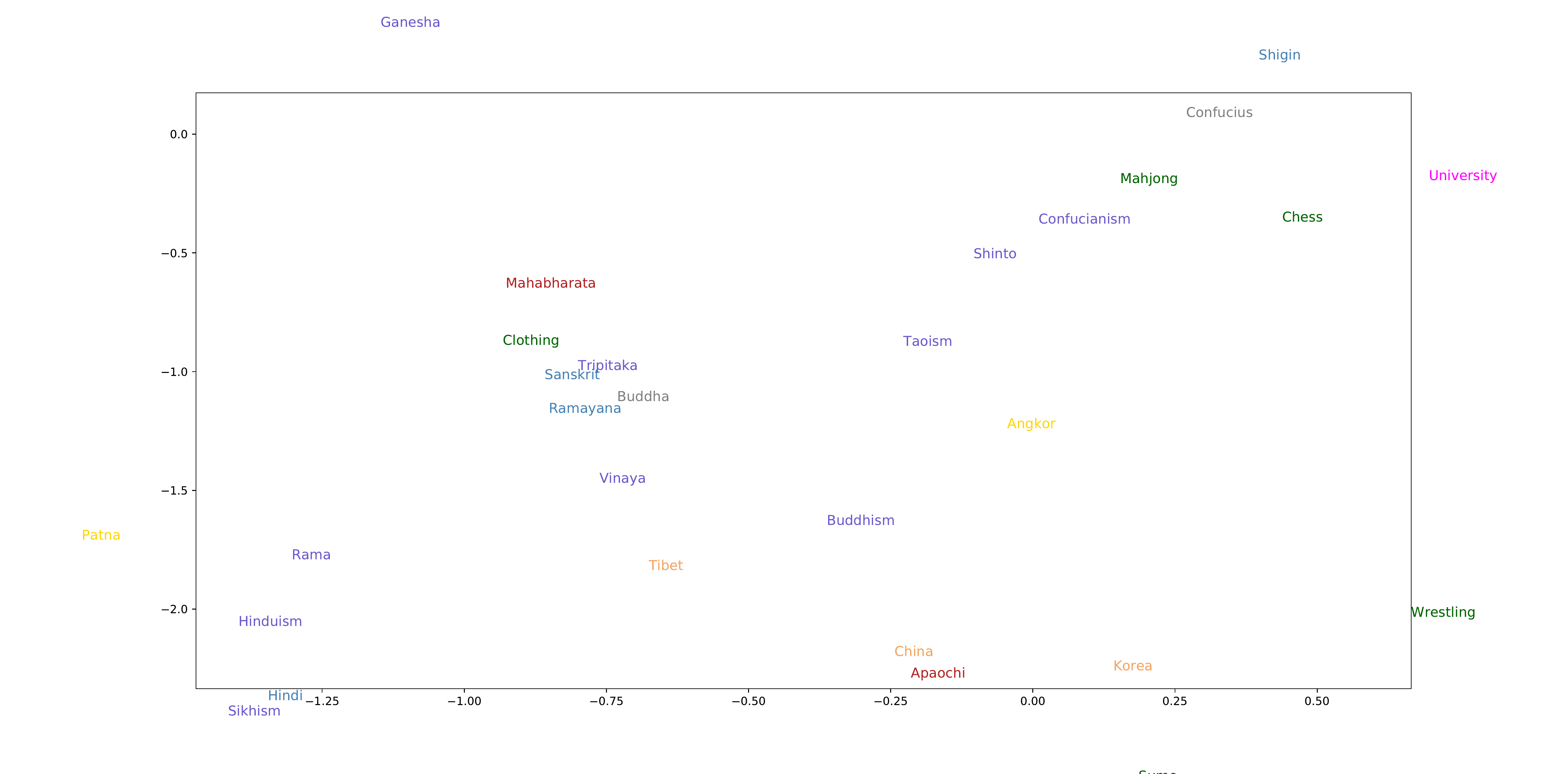}
\caption{Local plot of the ``red rectangle'' area in Figure 9(a).}\label{fig:visualize-local}
\end{figure}

From Figure \ref{fig:visualization} we can see that the node embeddings learned by different methods shows different characteristics in their distributions. For DeepWalk and Node2vec, the distributions are more intensive; for TransE and PTE the distributions are rather unconsolidated; and for GCN the distribution is streamline-like. The distribution of spread-gram is more balanced and discriminative compared with the other methods. However, we can still find that there is an area where nodes from multiple categories mixed (the area circled with red rectangle). To investigate the insights of these nodes, we enlarged this area and annotated the nodes with their corresponding entries, which is shown in Figure \ref{fig:visualize-local}.

From Figure \ref{fig:visualize-local} we can find that the entries are all related with religions or cultures of Southeast Asia. It implies that these entries, although belongs to different categories, constitute a meaningful semantic space. Therefore, the node embeddings of Wikipedia entries learned by spread-gram are semantically reasonable.

The above experimental results prove the effectiveness of spread-gram on network representation learning from multiple perspectives. We can conclude that spread-gram outperforms the baseline methods in almost all the tasks above.

\subsection{Iteration Analysis}

\begin{figure}[!htbp]
\centering
\subfloat [node classification in WITS]{\includegraphics[width=7cm]{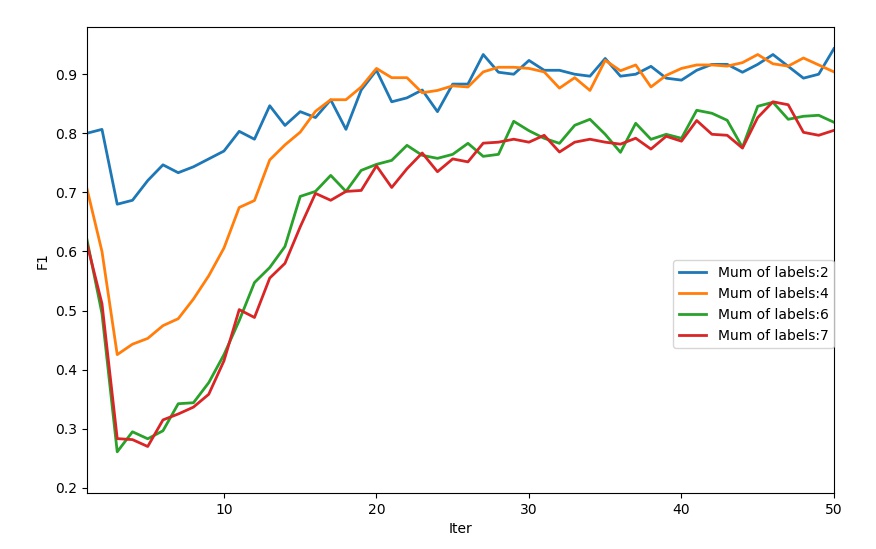}
\label{fig:iter-homo-a}}
\subfloat [link prediction in WITS]{\includegraphics[width=7cm]{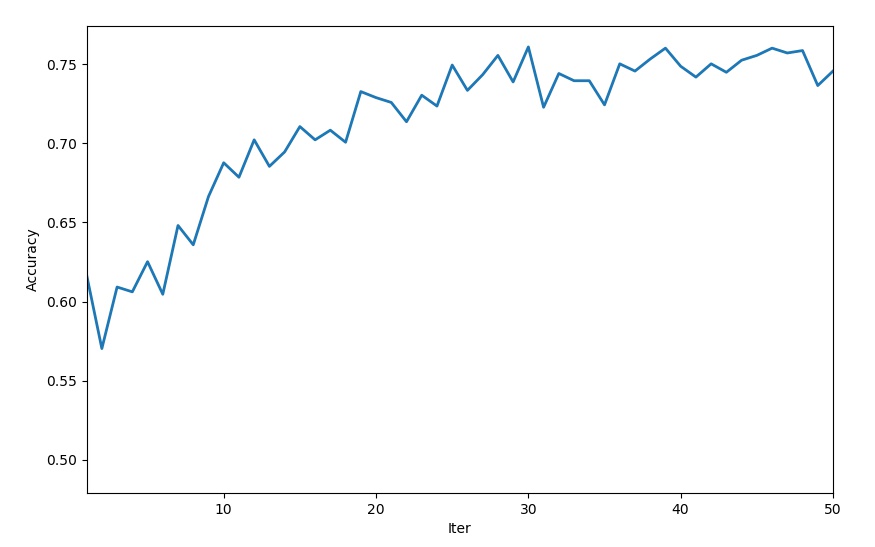}
\label{fig:iter-homo-b}}
\hfil
\subfloat [node classification in Wiki]{\includegraphics[width=7cm]{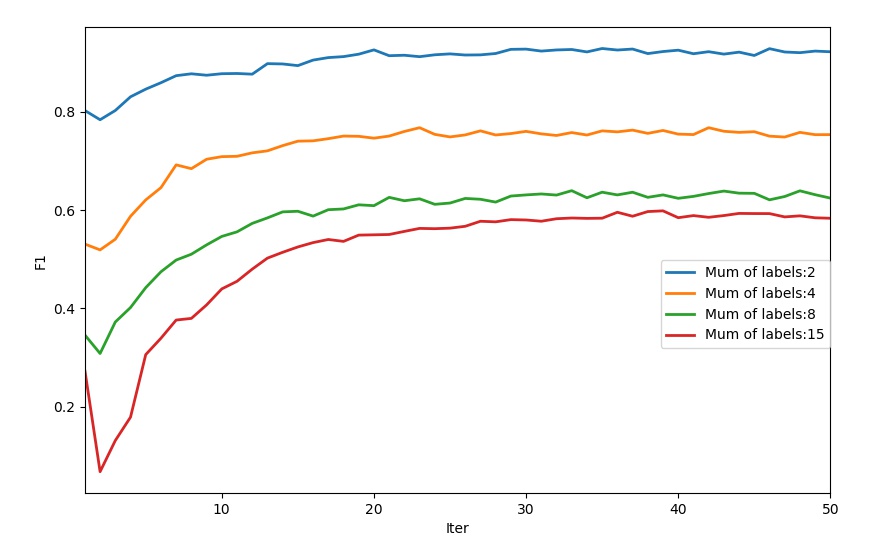}
\label{fig:iter-homo-c}}
\subfloat [link prediction in Wiki]{\includegraphics[width=7cm]{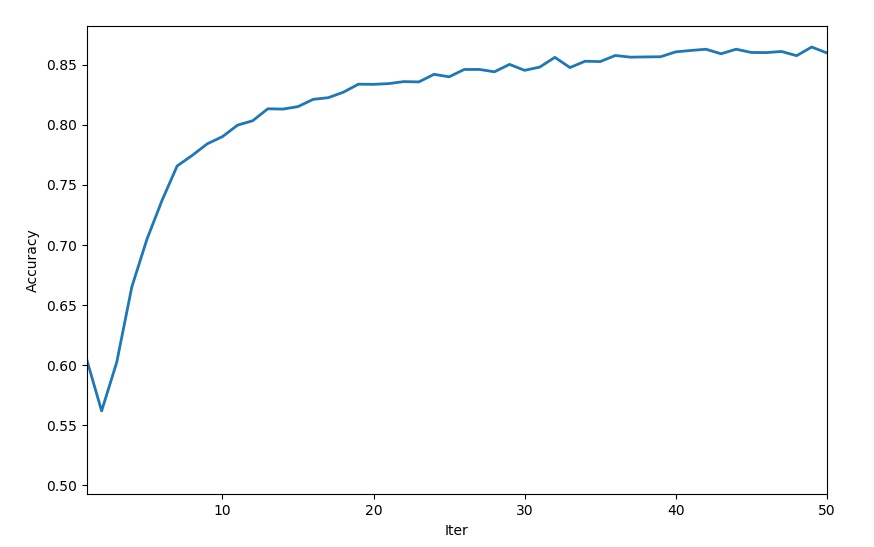}
\label{fig:iter-homo-d}}
\hfil
\subfloat [node classification in DIP]{\includegraphics[width=7cm]{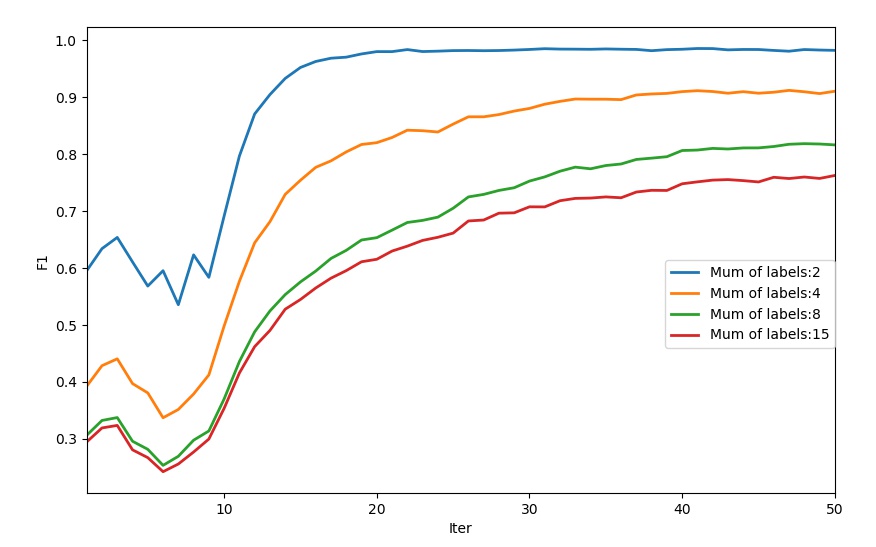}
\label{fig:iter-homo-e}}
\subfloat [link prediction in DIP]{\includegraphics[width=7cm]{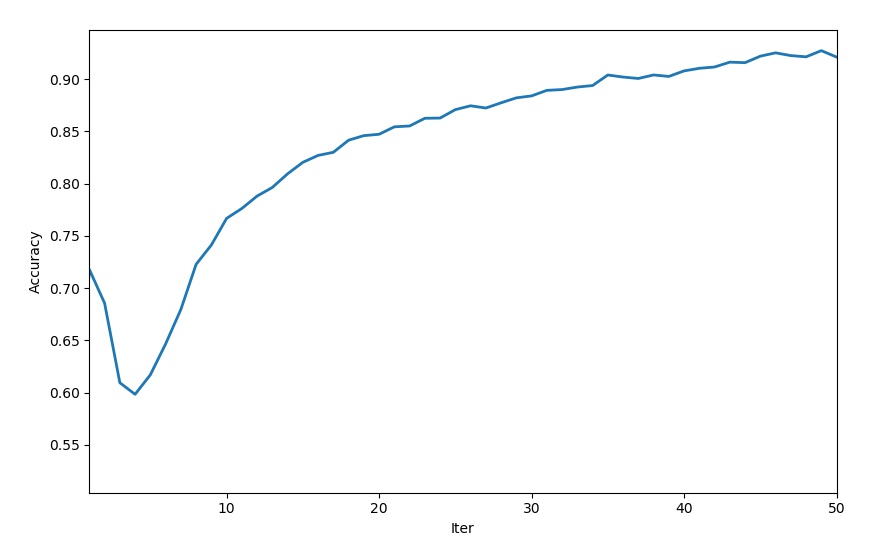}
\label{fig:iter-homo-f}}
\caption{The performance of spread-gram during the iterations (homogeneous networks).}
\label{fig:iter-homo}
\end{figure}

\begin{figure}[!htbp]
\centering
\subfloat [multi-classification in DBLP]{\includegraphics[width=7cm]{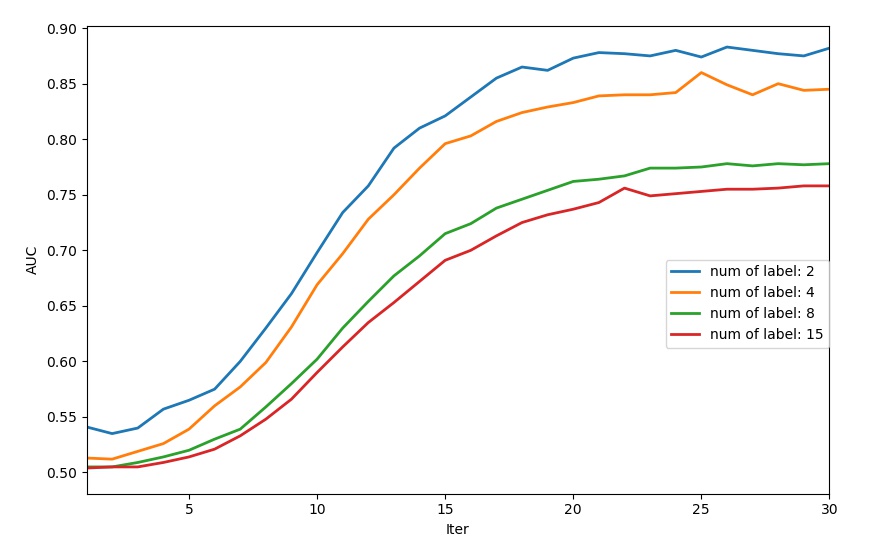}
\label{fig:iter-hetero-a}}
\subfloat [multi-classification in Amazon]{\includegraphics[width=7cm]{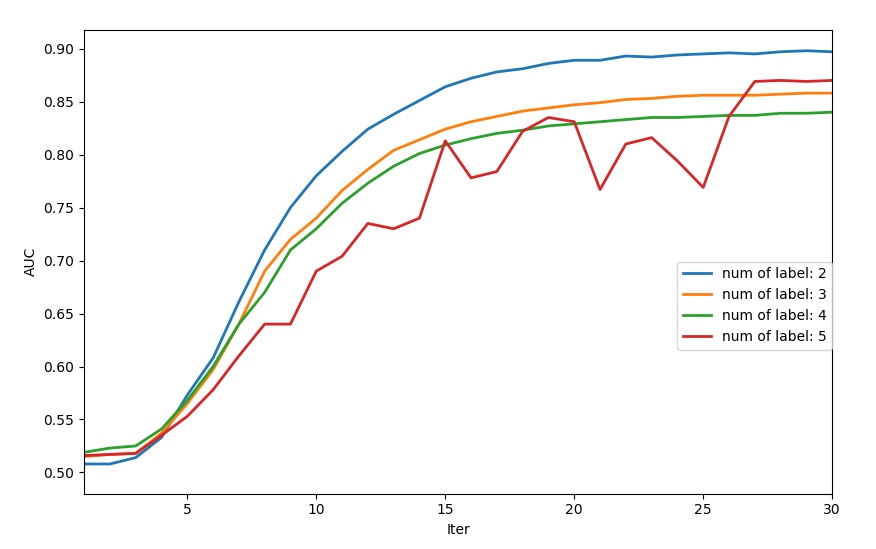}
\label{fig:iter-hetero-b}}
\hfil
\subfloat [node classification in DBLP]{\includegraphics[width=7cm]{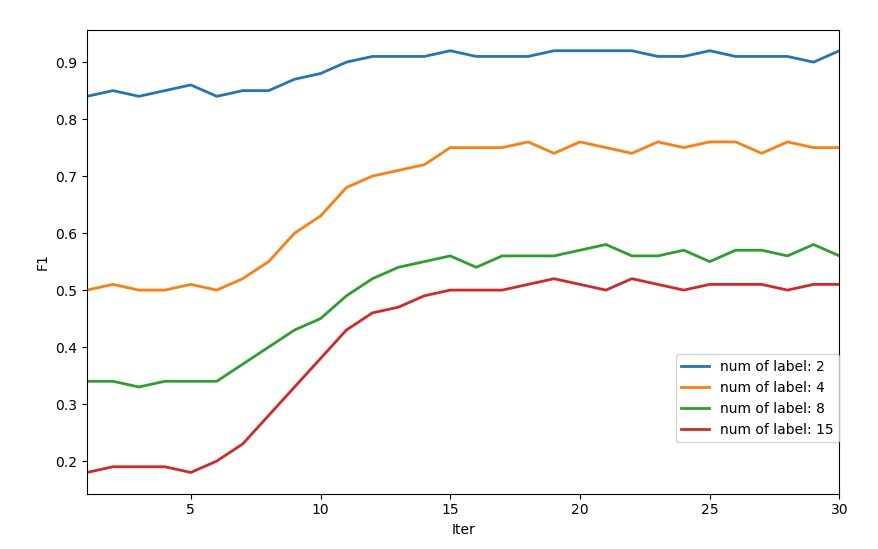}
\label{fig:iter-hetero-c}}
\subfloat [node classification in Amazon]{\includegraphics[width=7cm]{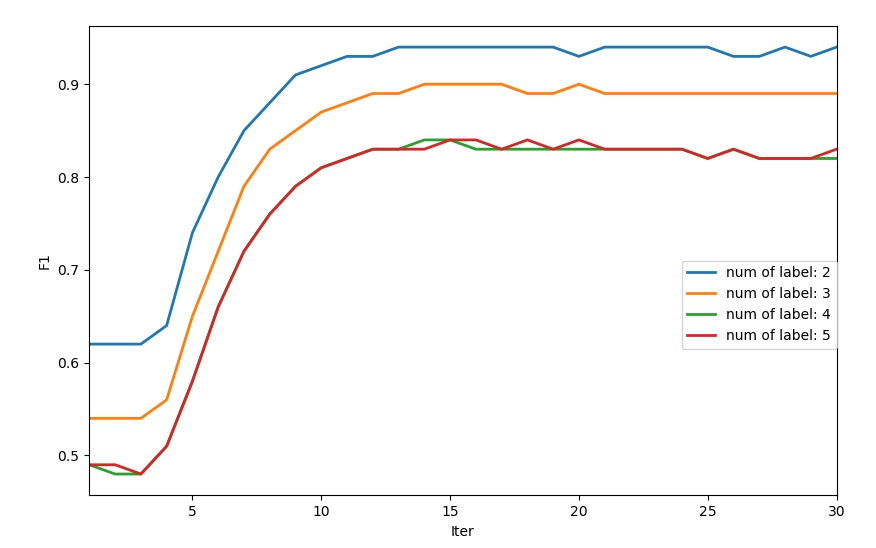}
\label{fig:iter-hetero-d}}
\hfil
\subfloat [link prediction in DBLP]{\includegraphics[width=7cm]{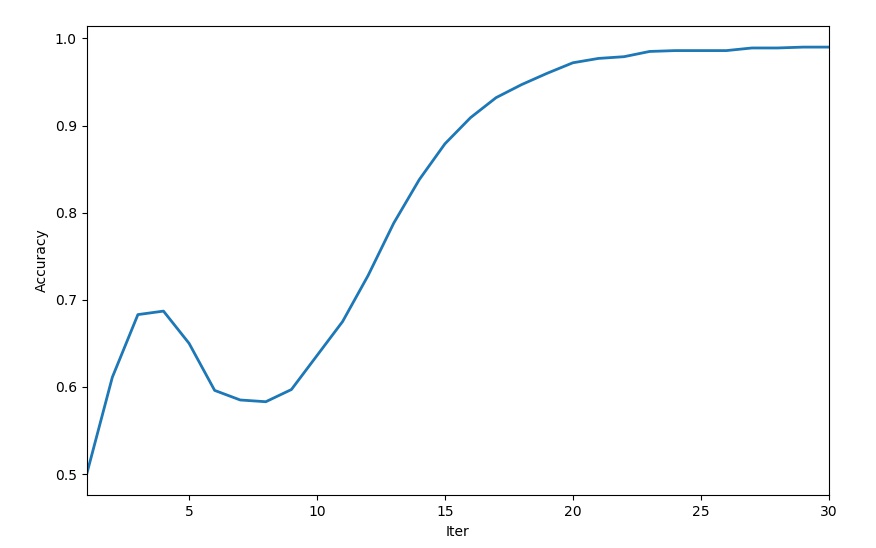}
\label{fig:iter-hetero-e}}
\subfloat [link prediction in Amazon]{\includegraphics[width=7cm]{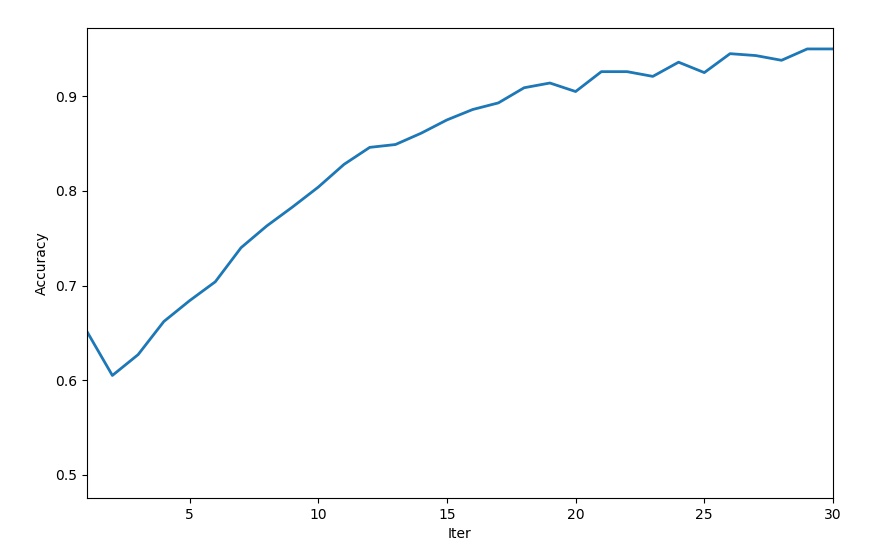}
\label{fig:iter-hetero-f}}
\caption{The performance of spread-gram during the iterations (heterogeneous networks).}
\label{fig:iter-hetero}
\end{figure}

To further analysis the performance of spread-gram during the iterations, we recorded the results of the above quantitative experiments for the spread-gram models after each iteration. More specifically, for different networks, we conducted the link prediction and node classification experiments once per iteration during the model training. The experimental settings follows the corresponding descriptions above. The results are shown in Figure \ref{fig:iter-homo} and Figure \ref{fig:iter-hetero}.

From Figure \ref{fig:iter-homo} and Figure \ref{fig:iter-hetero}, we can see that the models usually stabilize within a small number of iterations. For the most of the node classification experiments, the results converged at around the 15th iteration. For the link prediction experiments, the results converged slower, but usually within 30 iterations. This may because the object function of the spread-gram model maximizes the link prediction accuracy potentially, and the link prediction results stabilize at last. However, compared with the baseline methods, such as random-walk based methods which usually need more than 30 walk length, spread-gram converged with less computation cost.

%Reviews
\section{Related Works}
\label{Reviews}

Network representation learning attracted widely attention from the researchers during recent years. According to the techniques adopted, the network representation learning methods can be mainly divided into factorization based, random-walk based and neural-network based branches \cite{goyal2018graph}.

Factorization based methods were the mainstream in the early researches of network representation learning. By taking the network as an adjacent matrix, these studies usually get the lower dimension representation of the nodes through matrix factorization. The representative works include LLE \cite{roweis2000nonlinear}, IsoMap \cite{tenenbaum2000global}, Graph Factorization \cite{ahmed2013distributed}, EdgeCluster \cite{tang2009scalable}, as well as other methods applying SVD or Laplacian eigenmaps \cite{ou2016asymmetric, belkin2002laplacian}. However, these methods usually have high computational complexity, which cannot be applied effectively to the modern large-scale networks.

Random-walk based methods, due to their efficiency and flexibility on the large-scale networks, is adopted extensively in practical network analysis works. Deepwalk \cite{perozzi2014deepwalk} first combined random walk and word embedding learning models to get network representations. More specifically, it generates a number of node sequences through random walks on the network, and feed the node sequences to the word embedding learning model skip-gram \cite{mikolov2013efficient} to learn node representations. Afterwards, a series of models were proposed to extend the random-walk based methods, such as Node2vec \cite{grover2016node2vec}, metapath2vec \cite{dong2017metapath2vec}, anonymous walk \cite{ivanov2018anonymous}, HiWalk \cite{bai2019hiwalk}, etc. In spite of the advantages, the random-walk process distorts the origin structure of the networks and involves the input bias inevitably since it tries to reorganize the nonlinear networks with linear node sequences.

Neural network based network representation learning is an emerging area with the advances of deep learning applied to various fields. Multiple neural network structures, like autoencoders \cite{cao2016deep,wang2016structural}, convolutional neural networks \cite{kipf2016semi,derr2018signed,xu2019graph,lee2018higher} and sequential neural networks \cite{liu2017semantic} have been used in building network representation learning frameworks. Among these works, the convolutional networks based methods are the most extensively studied. Kipf and Welling proposed Graph Convolutional Networks (GCN) \cite{kipf2016semi}, which leverages the convolutional operation on nodes to collect associations from neighbors and update node embeddings. After proposed, GCN was further extended to multiple variations, such as signed Graph Convolutional Networks \cite{derr2018signed}, Graph Wavelet Neural Network \cite{xu2019graph}, and High-order Graph Convolutional Networks \cite{lee2018higher}. The Graph Convolutional Networks solve the input bias problems through combining the global and local structural information of the networks with graph convolution. However, they usually compute the network as a matrix, and the dynamical updating of local nodes cannot spread to the global network in a single layer. Moreover, the number of layers in a Graph Convolutional Networks is usually limited \cite{li2018deeper}.

\subsection{Network Analysis Applying Spreading Activation Theory}

Spreading activation theories has been applied in network analysis studies from various perspectives, like trust inference, churn prediction, word of mouth analysis, recommendation, etc.

Ziegler and Lausen \cite{ziegler2004spreading} believed that compared with investigating the interactions among entities, knowing about the entities’ credibility is equally important. They proposed Appleseed, a spreading-activation based model for trust computing in the network. The built local group trust metrics and evaluated credibility of entities through a trust propagation method, which borrowed ideas from spreading activation theories. Based on this work, Ziegler \cite{ziegler2015models} studied the “distrust propagation” problem under the spreading activation framework.

Dasgupta et.al \cite{dasgupta2008social} found that, in Telecom Business Intelligence, a critical factor that influence churn is the choice from friends. They proposed a spreading-activation based churn prediction method to predict user behavior according to the user’s neighboring nodes in a social network. Different from the method above, Kusuma et.al \cite{kusuma2013combining} and Kim et.al \cite{kim2014improved} used spreading-activation-based methods to evaluate the attributes of social ties, and predict churners according to these attributes. Existing studies \cite{oskarsdottir2016comparative} proved that the churn prediction models usually perform better through combining non-relational classifiers and relational classifiers, enhanced by the spreading-activation mechanism.

Yang et.al \cite{yang2010study} proposed a spreading-activation based viral marketing scheme. It considered both the location and the number of initial spreaders during the viral market modeling, which is proved to be effective in viral marketing. Besides, Wang et.al \cite{wang2018entagrec++} applied spreading-activation mechanism to recommendation, Grobe-bolting et.al \cite{grosse2015generic} applied spreading-activation theories to analyze user preferences.

%CONCLUSION
\section{Conclusion}

The main contributions of this paper are three-fold. First, we involve the human cognitive mechanism in network representation learning and prove its effectiveness. Second, we propose a new network representation learning method, spread-gram, which leverages the spreading-activation to overcome the limitations of existing method in integrating global structure and local structure. Third, we designed the spread-gram model learning methods for both homogeneous networks and heterogeneous networks. We conducted comprehensive experiments to evaluate the performance of spread-gram, and the results shows its effectiveness, efficiency and applicability to a wide range of real-world networks. 
We found a significant advantage of spread-gram in learning and representing the hierarchical structure of networks. 

The future improvements of this work could proceed from two perspectives. For one thing, analyzing weighted networks is an important issue, therefore how to deal with weighted network under the framework of spread-gram should be considered. For another, although spread-gram is developed for general networks, we should still find the scenes where the method is particularly suitable, such as information propagation network, human interaction network, etc.

%\clearpage
%\newpage
%\nocite{*}

%BIBLIO
%\bibliographystyle{apa}

\bibliographystyle{ieeetr}
\bibliography{spreadgram.bib} 

\end{document}